\documentclass[runningheads]{llncs}

\usepackage[final,year=2026]{eccv}

\usepackage{eccvabbrv}

\usepackage{graphicx}
\usepackage{bm}
\usepackage{booktabs}
\usepackage{multirow}
\usepackage{pifont}
\usepackage{threeparttable}
\usepackage[accsupp]{axessibility}
\usepackage{flafter}

\usepackage{hyperref}

\def\eqref#1{equation~\ref{#1}}

\def\1{\bm{1}}

\def\rv{{\textnormal{v}}}

\def\rvg{{\mathbf{g}}}
\def\rvh{{\mathbf{h}}}

\def\rvk{{\mathbf{k}}}

\def\rvp{{\mathbf{p}}}
\def\rvq{{\mathbf{q}}}
\def\rvr{{\mathbf{r}}}

\def\rvv{{\mathbf{v}}}
\def\rvw{{\mathbf{w}}}
\def\rvx{{\mathbf{x}}}
\def\rvy{{\mathbf{y}}}
\def\rvz{{\mathbf{z}}}

\def\rmE{{\mathbf{E}}}

\DeclareMathAlphabet{\mathsfit}{\encodingdefault}{\sfdefault}{m}{sl}
\SetMathAlphabet{\mathsfit}{bold}{\encodingdefault}{\sfdefault}{bx}{n}

\def\sR{{\mathbb{R}}}

\newcommand{\cmark}{\ding{51}}
\newcommand{\xmark}{\ding{55}}

\begin{document}

\title{Efficient Generative Adversarial Networks Using Linear Additive-Attention Transformers}

\titlerunning{LadaGAN}

\author{Emilio Morales-Juarez\inst{1} \and
Gibran Fuentes-Pineda\inst{2}}
\authorrunning{E.~Morales-Juarez and G.~Fuentes-Pineda}
\institute{Facultad de Ingenier\'ia, Universidad Nacional Aut\'onoma de M\'exico, Mexico\\
\email{emilio.morales@fi.unam.edu}
\and
Instituto de Investigaciones en Matem\'aticas Aplicadas y en Sistemas, Universidad Nacional Aut\'onoma de M\'exico, Mexico\\
\email{gibranfp@unam.mx}}

\maketitle

\begin{abstract}
Although deep generative models such as Diffusion Models (DMs) and Generative Adversarial Networks (GANs) achieve remarkable image generation performance, they often rely on computationally expensive architectures that limit accessibility and increase training and inference costs.
We introduce LadaGAN, a hybrid Transformer--convolutional GAN built around Ladaformer, an efficient linear-attention Transformer block.
Its linear additive-attention mechanism computes a single attention vector per head, avoiding the quadratic complexity of dot-product attention.
By combining Ladaformer with convolutional layers in both the generator and discriminator, LadaGAN reduces computational complexity while improving the training stability of Transformer-based GANs.
LadaGAN matches or surpasses the compared convolutional and Transformer GANs on benchmark datasets while being substantially more efficient, and remains competitive with multi-step generative models using orders of magnitude fewer computational resources.
While maintaining competitive FID scores, LadaGAN achieves over 600$\times$ higher inference throughput than ADM (333 vs.\ 0.50 images/s on CelebA $64\times64$) and requires over 100$\times$ fewer FLOPs.
Unlike one-step Consistency Models, LadaGAN can be trained efficiently on a single GPU.
\end{abstract}

\section{Introduction}
In recent years, deep generative models have achieved remarkable results in image generation. 
In particular, Generative Adversarial Networks (GANs) \cite{goodfellow2014generative} and Diffusion Models (DMs) \cite{ho2020denoising} have become the state-of-the-art approaches for this task.
GANs generate images in a single forward pass by learning to map a latent code to realistic samples, whereas diffusion models iteratively refine noise into images using learned denoising processes. Despite their success, GANs and DMs are often computationally expensive, typically requiring millions (and sometimes billions) of parameters and multiple high-end GPUs to train effectively \cite{huang2024gan,dhariwal2021diffusion}. Moreover, both paradigms involve extensive training iterations: GANs often require prolonged training, while diffusion models are even more costly due to the need to optimize multi-step denoising trajectories across many iterations. This computational burden poses a barrier to accessibility, reproducibility, and rapid experimentation, especially for researchers or developers without access to large-scale infrastructure.

Additionally, training GANs remains notoriously unstable. A large body of research has explored improved objectives (e.g., the Wasserstein loss \cite{arjovsky2017wasserstein}) and regularization methods (e.g., spectral normalization \cite{miyato2018spectral}) to mitigate divergence and mode collapse. Further, state-of-the-art GANs often require laborious engineering and sophisticated neural modules, as seen in convolution-based models like StyleGAN \cite{karras2019style,karras2020analyzing}, which are computationally demanding in terms of both FLOPs and parameters.
Since self-attention has been shown to effectively learn long-range dependencies \cite{dosovitskiy2020image}, different GAN architectures that incorporate Transformers \cite{vaswani2017attention} have been proposed. However, self-attention can make GAN training even more unstable \cite{lee2021vitgan}, and its $O(N^2)$ complexity results in high computational demands \cite{lee2021vitgan,zhang2021styleswin}. 

This paper presents LadaGAN, an efficient hybrid GAN architecture for image generation that combines a linear additive-attention Transformer (Ladaformer) with convolutional layers.\footnote{The source code is available at \url{https://github.com/milmor/LadaGAN}.}
We employ Ladaformer combined with convolutional layers in both the generator and the discriminator of LadaGAN, allowing efficient processing of long sequences in both networks.
In the generator, this block progressively generates a global image structure from the latent space using attention maps. In the discriminator, the Ladaformer generates attention maps to distinguish real and fake images.
Notably, the design of LadaGAN reduces the computational complexity and overcomes the training instabilities often associated with Transformer GANs.

The main contributions of this paper are as follows:
\begin{itemize}
    \item \textbf{Ladaformer: linear additive attention for stable adversarial training.} We introduce Ladaformer, a Transformer block with linear additive attention that enables efficient long-range modeling while remaining stable under adversarial settings. Unlike standard attention, it avoids mode collapse and gradient instabilities common in GANs. Ladaformer is simple, interpretable, and does not require custom kernels or training tricks. We further benchmark multiple $O(N)$ attention mechanisms under the same low-resource setting, and find that Ladaformer consistently offers the best trade-off between quality and efficiency.
    
    \item \textbf{LadaGAN: a lightweight, stable hybrid Transformer-convolutional GAN.} LadaGAN is designed to enable training from scratch on a single GPU, with significantly reduced training time and computational cost. By integrating Ladaformer blocks with convolutional layers in both the generator and discriminator, the architecture achieves high efficiency, requiring far fewer FLOPs and parameters than diffusion-based models, consistency models, or conventional GANs.
    
    \item \textbf{Favorable FID--compute trade-off.} LadaGAN achieves competitive FID scores against the compared Transformer GANs, diffusion-based models, and CT on CIFAR-10, CelebA, FFHQ, and LSUN Bedroom—without distillation, transfer learning, or large-scale infrastructure. 
\end{itemize}

\section{Related Work}

Transformer-based GANs have shown competitive results compared to convolutional models such as BigGAN \cite{brock2018large} and StyleGANs \cite{karras2019style,karras2020analyzing}. TransGAN \cite{jiang2021transgan} employs gradient penalty \cite{arjovsky2017wasserstein,gulrajani2017improved} and grid self-attention to address quadratic complexity. ViTGAN \cite{lee2021vitgan} generates patches to reduce sequence length and uses L2 attention \cite{kim2021lipschitz} with modified spectral normalization \cite{miyato2018spectral} for stability. However, both require multiple GPUs for training (TransGAN: 16 V100 GPUs, ViTGAN: 1 TPU).

Due to transformer discriminator instability \cite{lee2021vitgan}, recent works use conv-based discriminators with transformers only in generators. HiT \cite{zhai2022scaling} uses multi-axis blocked self-attention, while StyleSwin \cite{zhang2021styleswin} employs SwinTransformer blocks. Nevertheless, these architectures require multiple GPUs for training (StyleSwin: 8 V100 GPUs, HiT: 1 TPU) and do not leverage transformers in discriminators.

In contrast, GANsformer \cite{hudson2021gansformer} combines self-attention and convolutions but only partially leverages transformers as a StyleGAN generalization. LadaGAN also combines convolutions and self-attention, but unlike GANsformer, uses additive attention instead of dot-product attention in both generator and discriminator to address quadratic complexity and training instability. 

Diffusion models \cite{song2020denoising} have outperformed GANs in several tasks \cite{nichol2021improved,dhariwal2021diffusion}, learning to reverse a multi-step noising process where each step requires a full forward pass. 
However, DDPM \cite{song2020denoising} and ADM \cite{dhariwal2021diffusion} are even more complex than GANs in parameters and FLOPs, and require multiple forward passes for generation. 
To address this limitation, Consistency Models (CT) \cite{song2023consistency} reduce inference to 2 steps, but despite being a one-step generation model, CT requires substantially more training samples than LadaGAN across resolutions (e.g., 114x more at $256 \times 256$: 2048M vs 18M), making single-GPU training impractical.

Recently, R3GAN \cite{huang2024gan} proposed a new GAN loss function and architecture to improve the training stability of GANs. 
However, R3GAN requires 8 GPUs for training and exhibits dramatically higher computational costs than LadaGAN: on CIFAR-10, R3GAN requires 7.021 GFLOPS for the generator and 7.037 GFLOPS for the discriminator, whereas LadaGAN requires only 0.757 GFLOPS and 0.700 GFLOPS respectively, representing a nearly 10x reduction in FLOPs. 
Moreover, R3GAN requires 250M training images (3.7x more) compared to LadaGAN's 68M, yet both report similar FID scores under different training protocols, underscoring LadaGAN's efficiency and single-GPU training capability.
\section{Method}
In this section, we introduce the proposed LadaGAN architecture. The key component of LadaGAN is a linear additive attention Transformer which is combined with convolutional layers to build the generator and discriminator blocks. To the best of our knowledge, this is among the first GANs that use linear additive attention with convolutional layers in both the generator and the discriminator. Note that the design of a GAN architecture with a Transformer discriminator has proven to be challenging due to the computing cost of the dot-product attention and the training instabilities associated with the gradient penalty \cite{lee2021vitgan} commonly used in GANs.

\subsection{Linear additive attention (Lada)}
The LadaGAN attention mechanism is inspired by Fastformer's \cite{wu2021fastformer} additive attention \footnote{Not to be confused with Bahdanau's attention \cite{bahdanau2014neural}}. 
This efficient $O(N)$ Transformer architecture was originally designed for text processing, achieving comparable long-text modeling performance to the original dot-product attention at a fraction of the computational cost. Instead of computing the pairwise interactions among the input sequence vectors, Fastformer's additive attention creates a global vector summarizing the entire sequence using a single attention vector computed from the queries. 

More specifically, this linear additive attention computes each weight by projecting the corresponding query vector $\rvq_i \in \mathbb{R}^d$ with a vector $\rvw \in\mathbb{R}^d$, i.e.:

\begin{equation}
\alpha_i=\frac{\exp(\rvw^T\rvq_i/\sqrt{d})}{\sum_{j=1}^N\exp(\rvw^T\rvq_j/\sqrt{d})}.
\end{equation}

\noindent where $d$ is the head dimension.

To model interactions, a global vector is computed as follows:

\begin{equation}
\rvg =\sum_{i=1}^N\alpha_i \rvq_i.
\end{equation}

An element-wise operation is performed between $\rvg $ and each key vector $\rvk_i \in \mathbb{R}^d$ to propagate the learned information, obtaining a vector $\rvp_i \in \mathbb{R}^d$ such that

\begin{equation}
\rvp_i= \rvg \odot \rvk_i,
\label{element-wise}
\end{equation}

\noindent where the symbol $\odot$ denotes element-wise product. 

Unlike Fastformer \cite{wu2021fastformer}, LadaGAN's attention mechanism does not compute a global vector for the keys; instead, an element-wise operation is performed between each vector $\rvp_i$ and the corresponding value vector $\rv_i \in \mathbb{R}^d$. This operation enables propagating the information of the attention weights $\alpha_i, i = 1, \dots, N$ instead of compressing it. Finally, we compute each output vector $\rvr_i \in \mathbb{R}^d$ as 

\begin{equation}
\rvr_i = \rvp_i \odot \rvv_i. 
\end{equation}

\subsection{Ladaformer}

\begin{figure}[tb]
    \centering
    \includegraphics[width=0.55\textwidth]{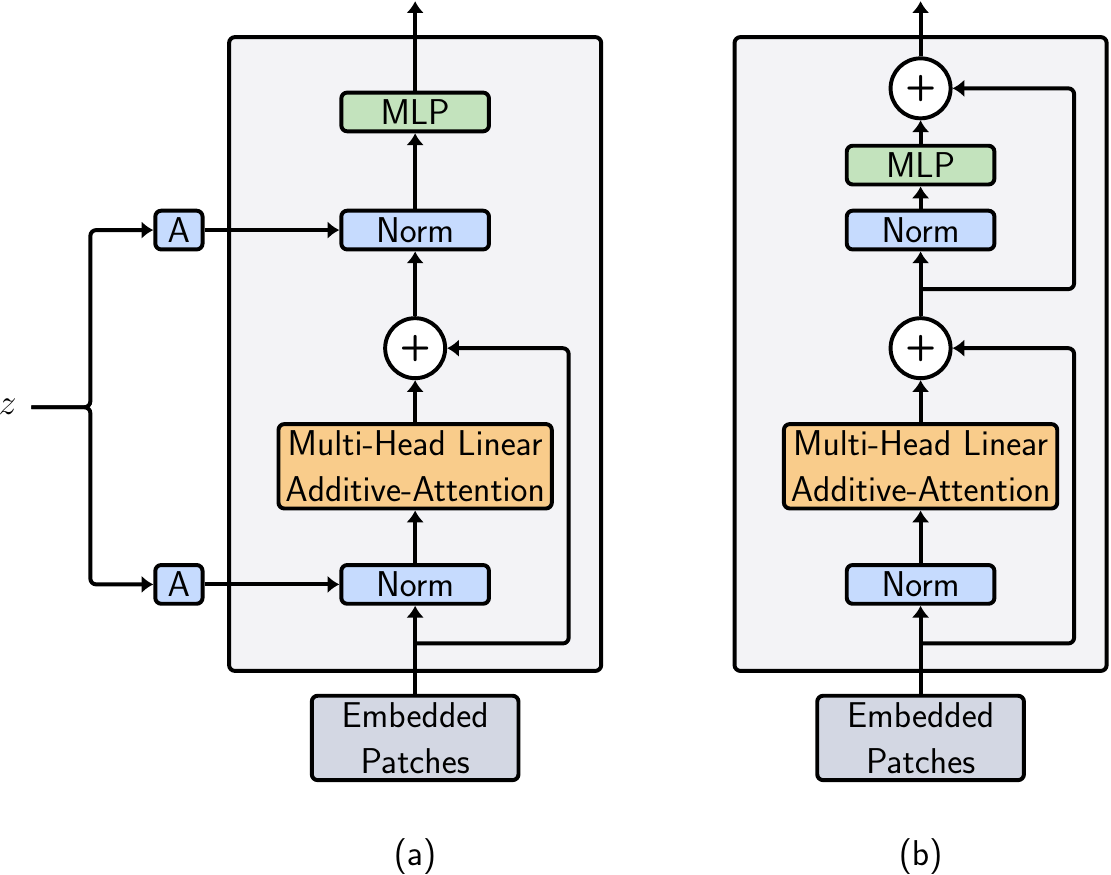}
    \caption{Ladaformer: (a) generator with SLN and without MLP residual connection and (b) discriminator without SLN and with MLP residual connection.}
    \label{fig:adat_trans}
\end{figure}

The main block of the generator and discriminator is Ladaformer, which closely follows the Vision Transformer (ViT) architecture \cite{dosovitskiy2020image}, as illustrated in Figure \ref{fig:adat_trans}. However, since introducing self-modulation has shown to be an effective strategy to improve performance \cite{chen2018selfmod,lee2021vitgan}, the LadaGAN generator block uses self-modulated layer normalization instead of standard layer normalization. 
In particular, layer normalization parameters for the inputs $\mathbf{h}_\ell$ of the $\ell$-th layer are adapted by 

\begin{equation}
    \text{SLN}(\rvh_\ell, \rvz) = \bm\gamma_\ell(\rvz) \odot \left(\frac{\rvh_\ell - \bm\mu}{\mathbf{\bm\sigma}}\right)  + \bm\beta_\ell(\rvz),
\end{equation}

\noindent where the division operation is performed element-wise. 

Note that this is slightly different from ViTGAN's self-modulated layer normalization, which injects a vector $\rvw$ computed by passing the latent vector $\rvz$ through a projection network; in contrast, LadaGAN injects $\rvz$ directly.
In addition, unlike ViT, ViTGAN, and Fastformer, the LadaGAN generator does not have the residual connection from the output of the attention module to the output of the multi-layer perceptron (MLP). 

\begin{align}
\mathbf{h^\prime}_\ell &= \text{MAA}(\text{SLN}(\mathbf{h}_{\ell-1}, \rvz)) + \mathbf{h}_{\ell-1},   \\
    \mathbf{h}_\ell &= \text{MLP}(\text{SLN}(\mathbf{h^\prime_{\ell}, \rvz})),
\end{align}

\noindent where $\text{MAA}(\cdot)$ denotes the multi-head linear additive attention and $\text{MLP}(\cdot)$ is a two-layer fully connected network with a GELU activation function in the first layer. 

\subsection{Generator}
\begin{figure}[tb]
    \centering
    \includegraphics[width=0.54\textwidth]{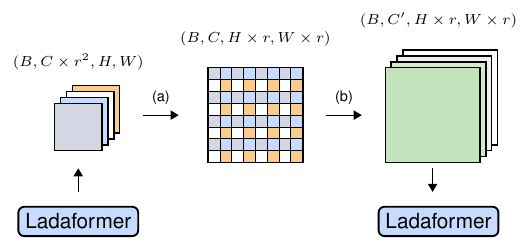}
\caption{Local Embedding Expansion. The output of the first Transformer consists of 4 feature maps that are expanded using pixel shuffle \textbf{(a)} to generate a single feature map. Then, the output map from pixel shuffle passes through a convolutional layer \textbf{(b)} to expand to 4 channels. These new 4 feature maps are the input to the second Transformer. In this way, even though the Transformers process sequences of different lengths, the dimensions of the embeddings are independent of the Pixel Shuffle operation.}
    \label{fig:expand}
\end{figure}

The LadaGAN generator employs the pixel shuffle operation to progressively transform the latent vector into an image. This operation is a common technique to increase spatial resolution, in which the input is reshaped from $(B, C \times r^2, H, W)$ to $(B, C, H \times r, W \times r)$, where $r$ is a scaling factor, $B$ is the batch size, $C$ is the number of channels of the output and $H$ and $W$ are the height and width of the input. Although this technique was originally proposed as an efficient alternative to standard ConvNet-based upsampling in super-resolution architectures, it has been widely adopted in image generation Transformers, including recent Transformer GANs. 
Since a pixel shuffle operation reduces the number of channels $C$ in the input (process (a) in Figure ~\ref{fig:expand}), we apply a convolutional layer after such an operation to increase the number of channels; we denote this operation as Local Expansion of the Embedding 

\begin{equation}
\text{LEE}(\mathbf{h}_\ell) = \text{Conv}(\text{PixelShuffle}(\mathbf{h}_\ell))
\end{equation}
\noindent where $\text{Conv}(\cdot)$ is a standard convolutional layer with $K$ filters and $\text{PixelShuffle}(\cdot)$ denotes the pixel shuffle operation with $r = 2$.  

\begin{figure}[tb]
    \centering
    \begin{subfigure}[t]{\textwidth}
    \centering
       \includegraphics[width=\textwidth]{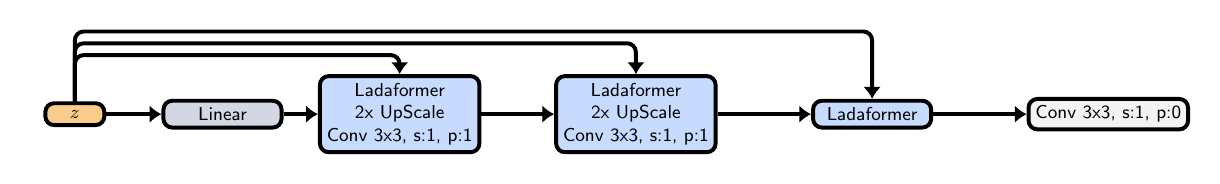}
      \caption{}
    \end{subfigure}
    \begin{subfigure}[t]{\textwidth}
\centering
   \includegraphics[width=\textwidth]{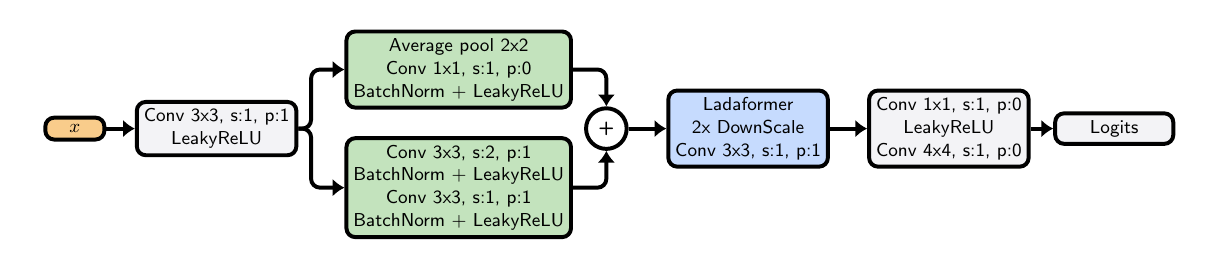}
  \caption{}
\end{subfigure}
    \caption{LadaGAN architecture: Lada-Generator (a) and Lada-Discriminator (b).}
    \label{fig:models}
\end{figure}

The LadaGAN generator uses the same architecture for the resolutions $32 \times 32$, $64 \times 64$, $128 \times 128$, and  $256 \times 256$, which consists of three Ladaformer blocks, as shown in Figure \ref{fig:models} (a). 
Since increasing the sequence length of Transformer models has generally improved performance in natural language processing tasks, we posit that similar benefits can be obtained in image generation tasks. Therefore, taking advantage of the $O(N)$ complexity of Ladaformer blocks, we aim to generate a long sequence (1024 tokens) as the output of the final Transformer block.

\begin{figure}[tb]
    \centering
    \includegraphics[width=0.54\textwidth]{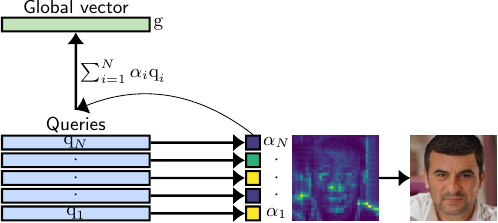}
\caption{Linear Additive Attention mechanism of a single head, generating a $32 \times 32$ map to construct a global structure for the image. This process guides the generation of patches for $128 \times 128$ image resolution.}
    \label{fig:attn}
\end{figure}

Given the latent vector $\rvz \in \sR^{D_{\rvz}}$, and $L$ Transformer blocks, LadaGAN generator operates as follows:

\begin{align}
    \rvh_0 &= \text{Linear}(\rvz),  \label{eq:generator_embedding} \\
    \mathbf{h^\prime}_\ell &= \text{MAA}(\text{SLN}(\mathbf{h}_{\ell-1} + \rmE_{\ell-1}, \rvz)) + \mathbf{h}_{\ell-1},   \\
    \mathbf{h}_\ell &= \text{LEE}(\text{MLP}(\text{SLN}(\mathbf{h^\prime_{\ell}, \rvz}))), \label{eq:lle}\\
    \rvy &= \text{MAA}(\text{SLN}(\mathbf{h}_{L} + \rmE_{L}, \rvz)) + \mathbf{h}_{L}, \\
    \rvx &= \text{Conv}(\text{MLP}(\text{SLN}(\rvy, \rvz))), \label{eq:channel}
\end{align}

\noindent where $\ell=1,\ldots,L$, $\text{Linear}(\cdot)$ denotes a linear projection, $\rmE_{L}  \in \mathbb{R}^{N_L \times D_L}$ and $\rmE_{\ell-1} \in \mathbb{R}^{N_{\ell-1} \times D_{\ell-1}} $ are the positional embeddings for the blocks $L$ and $\ell-1$ respectively, and $\rvx \in \sR^{H \times W \times C} $ is the output image. Note that before the final convolutional layer in equation \ref{eq:channel} and every pixel shuffle operation, a reshape operation is performed to generate a 2D feature map. On the other hand, if the number of output channels of the LEE convolution in equation \ref{eq:lle} is equal to the number of input channels, there is no expansion of the embedding dimension. This results in the convolution only reinforcing the pixel shuffle locality. Figure \ref{fig:attn} shows the LadaGAN generative process.

\subsection{Discriminator} 

LadaGAN discriminator resembles the FastGAN \cite{liu2021towards} discriminator but uses a Ladaformer instead of a residual convolutional block; the architecture of the LadaGAN discriminator is illustrated in Figure \ref{fig:models} (b). 
In particular, we found that batch normalization \cite{ioffe2015batch} in the convolutional feature extractor proves to be essential to complement the stability of the Lada discriminator. Note that batch normalization is not typically employed by Transformer discriminators, such as ViTGAN and TransGAN. 

In contrast to the LadaGAN generator, the discriminator Ladaformer block has the standard MLP residual connection, as shown in Figure \ref{fig:adat_trans} (b). 
In addition, a $\text{SpaceToDepth}(\cdot)$ operation is performed at the output. As opposed to $\text{PixelShuffle}(\cdot)$, $\text{SpaceToDepth}(\cdot)$ down-samples the input by reshaping it from $(B, C, H \times r, W \times r)$ to $(B, C \times r^2, H, W)$.
Unlike the final layer of the TransGAN and ViT discriminators that uses the class embedding, the final layer of the LadaGAN discriminator consists of convolutions with strides of $2$. In this way, the convolutions progressively reduce the attention map representation.

\subsection{Loss function}
LadaGAN employs standard non-saturating logistic GAN loss with $R_1$ gradient penalty \cite{mescheder2018training}. The $R_1$ term penalizes the gradient on real data, allowing the model to converge to a good solution. For this reason, it has been widely adopted in state-of-the-art GANs with convolutional discriminators. More specifically, the loss function is defined as follows:
\begin{equation}
\begin{split}
\label{eq:R1_D}
\mathcal{L}_D = &- \mathbb{E}_{\rvx \sim P_{\rvx}}[\log(D(\rvx))] - \mathbb{E}_{\rvz \sim P_{\rvz}}[\log(1 - D(G(\rvz)))] \\
&+ \gamma \cdot \mathbb{E}_{\rvx \sim P_{\rvx}} [ (\|\nabla_{\rvx} D(\rvx) \|^2_2)], 
\end{split}  
\end{equation}

\begin{equation}
\begin{split}
\label{eq:R1_G}
\mathcal{L}_G = &-\mathbb{E}_{\rvz \sim P_{\rvz}}[\log(D(G(\rvz )))].
\end{split}  
\end{equation}

\section{Experiments}
We evaluate LadaGAN's performance, efficiency, and stability through ablation studies and comparisons with state-of-the-art models. 

\subsection{Experiment setup}
We evaluate LadaGAN on CIFAR-10 \cite{krizhevsky2009learning} ($32 \times 32$), CelebA \cite{liu2015faceattributes} ($64 \times 64$), FFHQ \cite{karras2020analyzing} ($128 \times 128$), and LSUN Bedroom \cite{yu15lsun} ($128 \times 128$ and $256 \times 256$). We report FID \cite{heusel2017gans} and Recall, following evaluation protocols from previous works for fair comparison.

\begin{table}[tb]
\centering
\caption{FID and number of FLOPs for the LadaGAN generator using Linformer, Swin-Transformer, Fastformer, Ladaformer, and Lada-Swin attention mechanisms on CIFAR-10 ($32 \times 32$).}
\label{config-table}
\small
\renewcommand{\arraystretch}{0.95}
\begin{threeparttable}
\begin{tabular}{l@{\hspace{0.5em}}c@{\hspace{0.3em}}c@{\hspace{0.3em}}c@{\hspace{0.3em}}r@{\hspace{0.5em}}r@{\hspace{0.3em}}r}
\toprule
\textbf{Attention} & \textbf{Conv} & \textbf{Res-MLP} & 
\textbf{Mod} & \textbf{G-FLOPs}  & \textbf{FID $\downarrow$} & \textbf{Rec $\uparrow$} \\
\midrule
Swin & \xmark & \cmark & \cmark & 0.5B & 7.96 & 0.57 \\
Lada-Swin & \xmark & \xmark & \cmark & 0.4B & 6.46 & 0.58 \\
\midrule
\multirow{3}{*}{Linformer} & \xmark & \cmark & \cmark & 0.6B & 5.94 & 0.58 \\
 & \cmark & \cmark & \cmark & 0.8B & 6.59 & \textbf{0.60} \\
 & \cmark & \xmark & \cmark & 0.8B & 10.65 & 0.49 \\
\midrule
\multirow{4}{*}{Fastformer} & \xmark & \cmark & \cmark & 0.4B & 6.60 & 0.59 \\
 & \cmark & \cmark & \cmark & 0.6B & 6.57 & 0.56 \\
 & \cmark & \xmark & \cmark & 0.6B & 5.27 & 0.59 \\
 & \cmark & \cmark & \cmark & 0.6B\tnote{a} & N/A & N/A \\
\midrule
\multirow{7}{*}{Ladaformer} & \xmark & \cmark & \cmark & 0.5B & 6.47 & 0.58 \\
 & \xmark & \xmark & \cmark & 0.5B & 5.88 & 0.56 \\
 & \cmark & \cmark & \cmark & 0.7B & 6.29 & 0.57 \\
 & \cmark & \xmark & \cmark & 0.7B & \textbf{4.82} & 0.59 \\
 & \cmark & \cmark & \xmark & 0.7B & 6.19 & 0.57 \\
 & \cmark & \xmark & \xmark & 0.7B & 5.87 & 0.53 \\
 & \cmark & \xmark & \cmark & 0.7B\tnote{b} & \textbf{3.29} & \textbf{0.60} \\
\bottomrule
\end{tabular}
  \begin{tablenotes}
\footnotesize
\item All models use a convolutional discriminator unless otherwise noted.
\item[a] Fastformer discriminator.
\item[b] Ladaformer discriminator.
\item ``N/A'' indicates that training consistently diverged across multiple runs.
    \end{tablenotes}
\end{threeparttable}
\end{table}

\subsection{Implementation details}\label{sec:setting}
We train with $R_1$ regularization \cite{mescheder2018training} and Adam optimizer \cite{kingma2014adam} ($\beta_1 = 0.5$, $\beta_2 = 0.99$). Generator learning rate is $0.0002$; discriminator learning rates are $0.0004$ (convolutional) and $0.0002$ (Transformer). Ladaformer blocks generate $8 \times 8$, $16 \times 16$, and $32 \times 32$ maps with dimensions $1024$, $256$, and $64$ respectively. We use 4 heads and MLP dimension $512$. For CIFAR-10 and FFHQ, we apply Translation, Color, and Cutout augmentation \cite{zhao2020differentiable} with bCR \cite{zhao2021improved} ($\lambda_{real} = 1.0$, $\lambda_{fake} = 0.1$).

\subsection{Ablation studies}
\label{sec:ablation_exp}
We compare Ladaformer with Linformer, Swin-Transformer, Fastformer\footnote{All Fastformer experiments were performed using the official implementation available at \url{https://github.com/wuch15/Fastformer}}, and Lada-Swin (Ladaformer with Swin-style down-sampling). All models are trained with comparable FLOPs using convolutional discriminators.

\begin{table}[tb]
\centering
  \caption{FID and number of FLOPs for ConvNet and Ladaformer discriminators with and without LEE on CIFAR-10 ($32 \times 32$), CelebA ($64 \times 64$) and LSUN Bedroom ($128 \times 128$).}
  \label{config-disc-table}
\small
\renewcommand{\arraystretch}{0.95}
  \begin{threeparttable}    
\begin{tabular}{l@{\hspace{0.5em}}l@{\hspace{0.3em}}c@{\hspace{0.3em}}r@{\hspace{0.3em}}r@{\hspace{0.3em}}r@{\hspace{0.3em}}r}
\toprule
\textbf{Dataset} & \textbf{D-type} & \textbf{LEE} & \textbf{G-FLOPs} & \textbf{D-FLOPs}  & \textbf{FID $\downarrow$} & \textbf{Rec $\uparrow$}   \\
\midrule
\multirow{4}{*}{CIFAR-10} & Conv\tnote{$\dagger$} & \xmark &  0.7B & 0.5B  &  4.72 & 0.57 \\
    & Conv\tnote{$\dagger$} & \cmark &  0.9B & 0.5B &  4.68 & 0.55\\
   & Lada\tnote{$\dagger$} & \xmark &  0.7B & 0.7B & \textbf{3.29} & \textbf{0.60}\\
     &Lada\tnote{$\dagger$} & \cmark & 0.9B & 0.7B & 3.60 & \textbf{0.60}\\
\midrule
\multirow{4}{*}{CelebA} & Conv & \xmark & 0.7B & 0.7B  &  3.43 & 0.55\\
     & Conv & \cmark & 0.9B & 0.7B  & 3.36 & 0.58\\
     & Lada & \xmark & 0.7B & 0.9B &  \textbf{2.89}  & 0.58\\
     & Lada & \cmark & 0.9B & 0.9B & 3.04 & \textbf{0.59}\\
\midrule
\multirow{4}{*}{\begin{tabular}{@{}c@{}}LSUN\\Bedroom\tnote{$\ast$}\end{tabular}} & Conv & \xmark & 0.9B & 0.9B  &  9.33 & 0.29\\
     & Conv & \cmark & 4.3B & 0.9B  &  5.82 & 0.36 \\
    & Lada & \xmark & 0.9B & 1.1B & 6.46  & 0.38 \\
     & Lada & \cmark & 4.3B & 1.1B & \textbf{4.60} & \textbf{0.41}\\
\bottomrule
  \end{tabular}
  \begin{tablenotes}
\footnotesize
  \item[$\ast$] Convolutional decoder with nearest neighbor upsampling instead of patch generation.
    \item[$\dagger$] With bCR regularization. 
    \end{tablenotes}
\end{threeparttable}
\end{table}

Table \ref{config-table} shows that Ladaformer with convolutions and without MLP residual connection achieves the best FID. 
Adding convolutions benefits Ladaformer but degrades Linformer's performance, suggesting that Ladaformer's element-wise attention propagation (equation \ref{element-wise}) complements convolutional locality better than dot-product attention. 
Ladaformer and Fastformer achieve lower FIDs without residual connections. In contrast, Linformer (dot-product attention) deteriorates significantly without residuals, indicating that linear additive attention is less dependent on this residual shortcut. 
The best configuration uses a Ladaformer discriminator; replacing it with Fastformer causes training divergence despite similar architecture.

\subsection{Lada discriminator} \label{sec:disc_exp}
\begin{figure}[tb]
  \centering
  \includegraphics[width=0.65\textwidth]{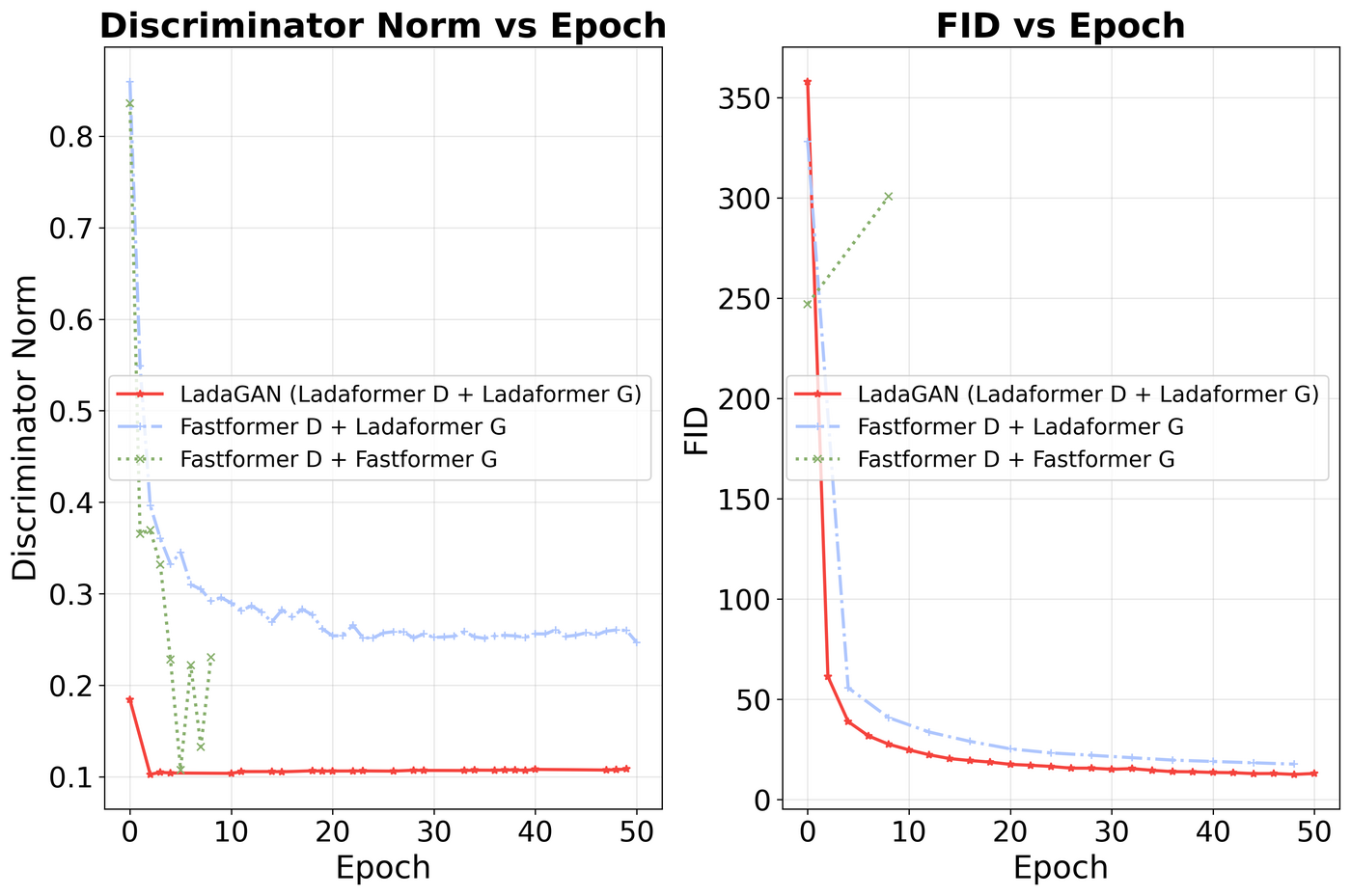}
  \caption{Comparison of Ladaformer GAN, Fastformer Discriminator + Ladaformer Generator, and Fastformer GAN on CIFAR-10: Discriminator Norm (left) and FID (right) vs training epochs.}
  \label{fig:fast_vs_lada}
\end{figure}

Transformer discriminators in GANs pose challenges because of $R_1$ gradient penalty instability \cite{lee2021vitgan}. 
We show that Ladaformer discriminators trained with $R_1$ consistently outperform convolutional discriminators across all datasets (Table \ref{config-disc-table}). 
The improvement is larger for LSUN Bedroom with bigger generators. 
LEE benefits LSUN Bedroom, a large-scale dataset, by effectively increasing the dimension of Ladaformers, which leads to improved FID performance. 
Figure \ref{fig:fast_vs_lada} illustrates that the Fastformer-based GAN suffered from divergence and failed to converge, in contrast to LadaGAN, which demonstrated stable training and superior performance.

\subsection{Data efficiency}
\begin{table}[tb]
\centering
\caption{FID scores for CIFAR-10 models trained using 100\%, 20\%, and 10\% of images, computed with 50k training images and 10k test images.}
\label{aug_table}
\small
\begin{threeparttable}
\begin{tabular}{l@{\hspace{0.5em}}r@{\hspace{0.3em}}r@{\hspace{0.3em}}r@{\hspace{0.3em}}r@{\hspace{0.3em}}r@{\hspace{0.3em}}r}
\toprule
\textbf{Method} &  \multicolumn{2}{c}{100\% data} & \multicolumn{2}{c}{20\% data} & 
\multicolumn{2}{c}{10\% data}\\
\cmidrule(lr){2-3}\cmidrule(lr){4-5}\cmidrule(lr){6-7}
 & 50k   & 10k    & 50k   & 10k & 50k   & 10k \\
\midrule
StyleGAN2\tnote{a}  &  \textsuperscript{$\ast$}5.79 & \textsuperscript{$\ast$}9.89   & --  & \textsuperscript{$\ast$}12.15  & -- & \textsuperscript{$\ast$}14.50\\
\midrule
LadaGAN  &  \textbf{3.29} & \textbf{7.58} & \textbf{6.85}  &  \textbf{11.06} & \textbf{8.93}  & \textbf{12.97}\\
w/o bCR &  4.88 & 9.01 &   10.67&  15.09  & 12.79  & 16.84 \\
\bottomrule
\end{tabular}
\begin{tablenotes}
\footnotesize
\item[a] \cite{karras2020analyzing}
\item[$\ast$] Results from \cite{zhao2020differentiable}.
\end{tablenotes}
\end{threeparttable}
\end{table}

We evaluate LadaGAN under reduced training set sizes on CIFAR-10 (Table~\ref{aug_table}).
Table \ref{aug_table} shows LadaGAN with bCR outperforms StyleGAN2 across all data regimes (10\%, 20\%, 100\%). Without bCR, performance declines in low-data regimes, highlighting bCR's importance for small datasets. 

\subsection{Comparison with state-of-the-art models} 
\label{sec:exp_i}
\begin{table}[tb]
\centering
\caption{Comparison with representative models. FID for CIFAR-10 with 50k samples, CelebA with 19k and 50k samples, FFHQ with 70k samples, and LSUN Bedroom with 30k and 50k samples. Except for SS-GAN, all Convolutional and Transformer GANs were trained using differentiable augmentation. \textsuperscript{$\ast$}Results from the original papers.}
\label{main-table}
\scriptsize
\begin{threeparttable}
\resizebox{\textwidth}{!}{%
\begin{tabular}{llrrrrrr}
\toprule
& \multirow{2}{*}{\textbf{Method}} & \textbf{CIFAR-10} &  \multicolumn{2}{c}{\textbf{CelebA}} & \textbf{FFHQ} & \textbf{LSUN} & \textbf{LSUN}\\
 & & \textbf{32$\times$32}& \multicolumn{2}{c}{\textbf{64$\times$64}} & \textbf{128$\times$128} & \textbf{128$\times$128} & \textbf{256$\times$256}\\
\cmidrule(lr){4-5}
 & & 50k & 19k & 50k & 50k & 30k & 50k \\
\midrule
\multirow{9}{*}{Single-step}
 & SS-GAN \cite{chen2019self} & \textsuperscript{$\ast$}15.60  & -- &-- & -- & \textsuperscript{$\ast$}13.30 & --\\
 & TransGAN \cite{jiang2021transgan} & \textsuperscript{$\ast$}9.02  & -- & -- & -- & -- & --\\
 & Vanilla-ViT \cite{lee2021vitgan} & \textsuperscript{$\ast$}12.70  & \textsuperscript{$\ast$}20.20  & -- & -- & -- & --\\
 & VITGAN \cite{lee2021vitgan} & \textsuperscript{$\ast$}4.92   & \textsuperscript{$\ast$}3.74  &  -- & -- & -- & --\\
 & GANformer \cite{hudson2021gansformer} & --   & -- &  -- & -- & -- & \textsuperscript{$\ast$}6.51\\
 & BigGAN + DiffAugment \cite{zhao2020differentiable} & \textsuperscript{$\ast$}4.61 & -- & -- & -- & -- & --\\
 & StyleGAN2 + DiffAugment \cite{zhao2020differentiable} & \textsuperscript{$\ast$}5.79 & -- & -- & -- & -- & --\\
 & StyleGAN2-D + ViTGAN-G \cite{lee2021vitgan}  & \textsuperscript{$\ast$}4.57 & -- & -- & -- & -- & -- \\
 & CT \cite{song2023consistency} & \textsuperscript{$\ast$}8.70 & -- & -- & -- & -- & \textsuperscript{$\ast$}16.0 \\
  & LadaGAN  & \textbf{3.29} & \textbf{2.89} & \textbf{1.81} & \textbf{4.48} & \textbf{5.08} & \textbf{6.36} \\
\midrule
\multirow{3}{*}{Multi-step}  &  CT (2 steps) \cite{song2023consistency} & \textsuperscript{$\ast$}5.83 & -- & -- & -- & -- & \textsuperscript{$\ast$}7.85 \\
    & ADM-IP (80 steps)\tnote{a} & \textsuperscript{$\ast$}2.93  & -- & \textsuperscript{$\ast$}2.67 & \textsuperscript{$\ast$}6.89 & -- & --  \\
    &  ADM-IP (1000 steps)\tnote{a} & \textsuperscript{$\ast$}\textbf{2.76}  & -- & \textsuperscript{$\ast$}\textbf{1.31} & \textsuperscript{$\ast$}\textbf{2.98}   & -- & -- \\
\bottomrule
\end{tabular}%
}
\begin{tablenotes}
\item[a] \cite{dhariwal2021diffusion,ning2023input}
\end{tablenotes}
\end{threeparttable}
\end{table}

\begin{table}[tb]
\centering
\caption{Computation cost for 80 ADM-IP steps, 2 CT steps, and samples seen (training iterations times batch size). Tput denotes inference throughput (images/s), measured on a single RTX 3090 for all reported models. For GANs, FLOPs are from the Generator. For CelebA ($64 \times 64$), LadaGAN is trained on a single NVIDIA 3080 Ti GPU in less than 35 hours, while ADM training takes 5 days on 16 Tesla V100 GPUs.}
  \label{flops-table}
\small
\renewcommand{\arraystretch}{0.88}
  \begin{threeparttable}    
\begin{tabular}{l@{\hspace{0.3em}}l@{\hspace{0.2em}}r@{\hspace{0.3em}}r@{\hspace{0.3em}}r@{\hspace{0.3em}}r@{\hspace{0.3em}}r}
\toprule
\textbf{Res.} & & CT  & ADM & StyleGAN2 & VITGAN & LadaGAN  \\
\midrule
\multirow{4}{*}{32$^2$} &  Params & 57M & 57M & -- &  -- & \textbf{19M} \\
    & FLOPs & 22B & 9.0B & \textsuperscript{$\dagger$}1.2B &  -- & \textbf{0.7B} \\
   & It.$\times$BS & 409M & 69M & -- &  -- &  \textbf{68M}  \\
  & Tput & -- & 0.66 & -- &  -- & \textbf{416.66} \\
\midrule
\multirow{4}{*}{64$^2$} &  Params & -- & 295M & \textsuperscript{$\ast$}24M &  \textsuperscript{$\ast$}38M & \textbf{19M} \\
    & FLOPs & -- & 103B & \textsuperscript{$\ast$}7.8B &  \textsuperscript{$\ast$}2.6B & \textbf{0.7B} \\
   & It.$\times$BS& -- & 138M & -- &  -- & \textbf{72M}   \\
   & Tput & -- & 0.50 & -- &  -- & \textbf{333.33}   \\
\midrule
\multirow{4}{*}{128$^2$} &  Params & -- & 543M & -- &  -- & \textbf{24M} \\
    & FLOPs & -- & 391B & \textsuperscript{$\ast$}11.5B &  \textsuperscript{$\ast$}11.8B & \textbf{4.3B} \\
   & It.$\times$BS & -- & 61M & -- &  -- & \textbf{53M}  \\
  & Tput & -- & 0.23 & -- &  -- & \textbf{192.30}  \\
\midrule
\multirow{4}{*}{256$^2$} &  Params & 526M & -- & -- &  -- & \textbf{24M} \\
    & FLOPs & -- & -- & \textsuperscript{$\ast$}15.2B &  \textsuperscript{$\ast$}52.1B & \textbf{5.5B} \\
   & It.$\times$BS & 2048M & -- & -- &  -- & \textbf{18M}  \\
  & Tput & 4.18 & -- & -- &  -- &  \textbf{63.69} \\
\bottomrule
  \end{tabular}
\begin{tablenotes}
\footnotesize
\item[$\ast$] Results from \cite{lee2021vitgan}.
\item[$\dagger$] StyleGAN2 $32 \times 32$ FLOPs from \cite{zhao2020differentiable}.
\end{tablenotes}
\end{threeparttable}
\end{table}

Table \ref{main-table} shows LadaGAN obtains the best FID among the compared single-step GANs and CT under a shared DiffAug protocol. Remarkably, LadaGAN remains competitive with ADM-IP (80 and 1000 steps) despite being single-step.
For fair architectural comparison, we use standard differentiable augmentation \cite{zhao2020differentiable} consistent with standard GAN evaluation protocols, matching the augmentation used by the models in our comparison table (StyleGAN2, BigGAN, ViTGAN, TransGAN). While some variants (e.g., StyleGAN2-ADA \cite{karras2020training}, R3GAN) employ more sophisticated augmentation strategies that yield better performance, we exclude them from our comparison as it would be unfair to compare models with different augmentation protocols. These advanced augmentation techniques are separate from architectural efficiency and can be integrated into LadaGAN in future work.

\begin{figure}[tb]
    \centering
    \begin{subfigure}[t]{0.44\textwidth}
      \includegraphics[width=\textwidth]{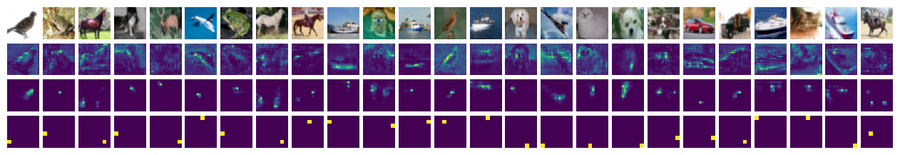}
      \caption{}
      \label{fig:adat_maps:a}
    \end{subfigure}
    \hfill
    \begin{subfigure}[t]{0.44\textwidth}
      \includegraphics[width=\textwidth]{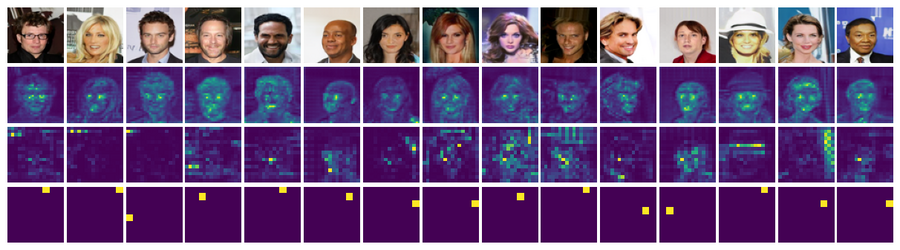}
      \caption{}
      \label{fig:adat_maps:b}
    \end{subfigure}
    \begin{subfigure}[t]{0.44\textwidth}
      \includegraphics[width=\textwidth]{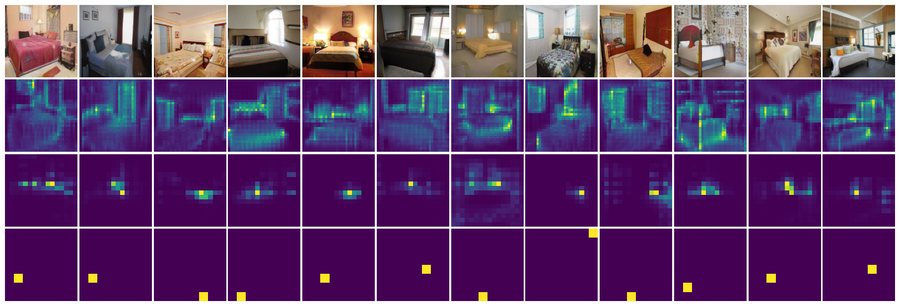}
      \caption{}
      \label{fig:adat_maps:c}
    \end{subfigure}
    \hfill
    \begin{subfigure}[t]{0.44\textwidth}
      \includegraphics[width=\textwidth]{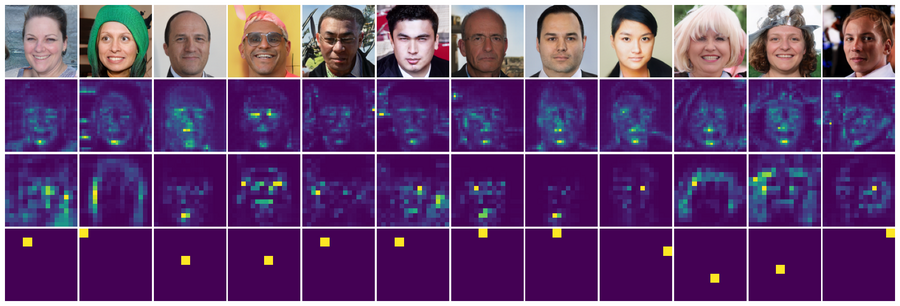}
      \caption{}
      \label{fig:adat_maps:d}
    \end{subfigure}
    \caption{Samples from LadaGAN on CIFAR-10 (a), CelebA (b), LSUN Bedroom (c), and FFHQ (d), with additive attention maps from three Ladaformer blocks.}
    \label{fig:adat_maps}
\end{figure}

\begin{figure}[tb]
    \centering
    \begin{subfigure}[t]{0.44\textwidth}
      \includegraphics[width=\textwidth]{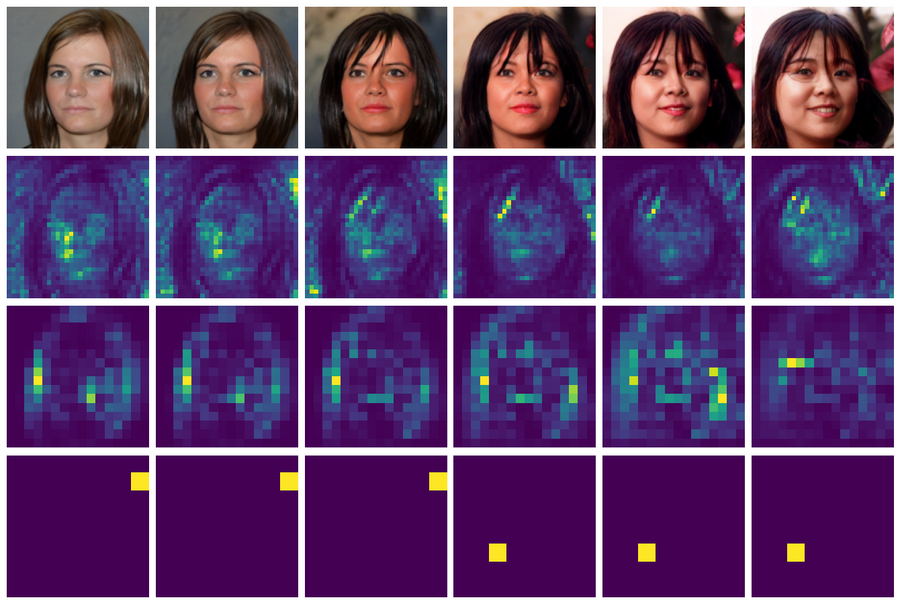}
      \caption{}
      \label{fig:inter_maps:a}
    \end{subfigure}
    \hfill
    \begin{subfigure}[t]{0.44\textwidth}
      \includegraphics[width=\textwidth]{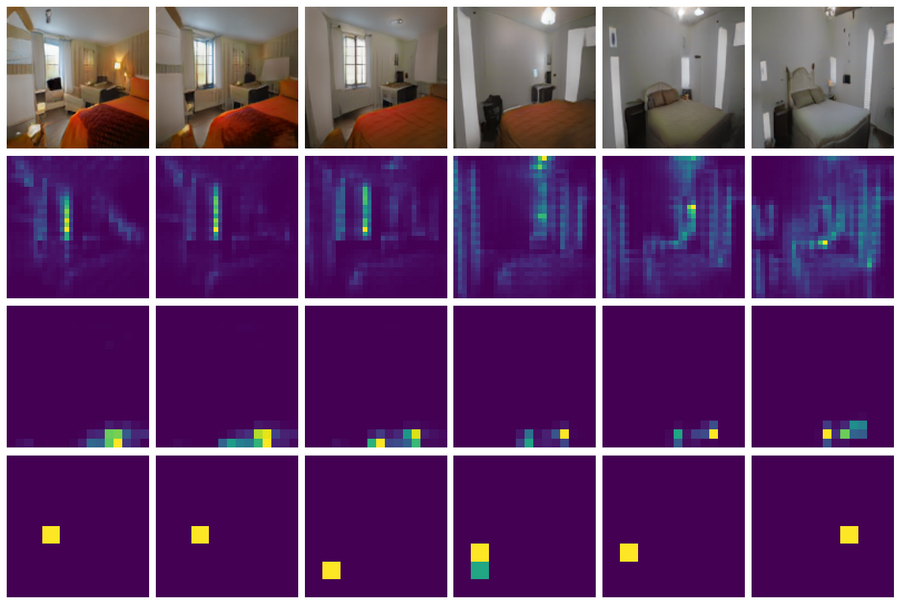}
      \caption{}
      \label{fig:inter_maps:b}
    \end{subfigure}
    \caption{Latent-space interpolations and attention maps from LadaGAN on FFHQ (a) and LSUN Bedroom (b).}
    \label{fig:inter_maps}
\end{figure}

Figure~\ref{fig:adat_maps} shows generated samples with additive attention maps from LadaGAN. Figure~\ref{fig:inter_maps} shows latent-space interpolations on FFHQ and LSUN Bedroom.

\subsection{Computational cost analysis}
Table \ref{flops-table} summarizes computation cost across resolutions; LadaGAN requires the fewest parameters and FLOPs, with parameter count nearly constant between $32 \times 32$ and $64 \times 64$ due to patch generation ($2 \times 2$).

\subsubsection{Comparison with GANs}
LadaGAN requires only 8.9\% of StyleGAN2 and 26.9\% of ViTGAN parameters at $64 \times 64$. At $256 \times 256$, Table \ref{flops-table} shows StyleGAN2 requires 15.2B FLOPs, ViTGAN requires 52.1B FLOPs, and LadaGAN requires only 5.5B FLOPs. At $256 \times 256$, StyleGAN2 requires 13 days to process 10M images (it$\times$BS) on NVIDIA DGX-1 using a single Tesla V100 GPU \cite{karras2020analyzing}, or 1 day 22 hours with 8 GPUs, whereas LadaGAN processes 10M images in 25 hours on a single RTX 3090, demonstrating the efficiency gains from reduced FLOPs. Given that ViTGAN requires 9.5x more FLOPs than LadaGAN at $256 \times 256$, single-GPU training is not viable for ViTGAN.

\subsubsection{Comparison with diffusion models and CT}
Multi-step models (ADM, CT) require orders of magnitude more FLOPs than GANs. 
Remarkably, ADM CelebA ($64 \times 64$) requires 5 days on 16 Tesla V100 GPUs \cite{ning2023input}, while LadaGAN trains in 35 hours on a single RTX 3080 Ti.
On CelebA $64 \times 64$, LadaGAN achieves over 600x higher inference throughput on the same RTX 3090 (333 images generated/s vs 0.50 images generated/s) and uses over 100x fewer FLOPs compared to ADM. 
To match the inference speed of LadaGAN, a diffusion model would need to be distilled to a single step, which requires additional training and typically degrades quality significantly. 
Notably, despite CT being a one-step generation model, it requires 114x more training samples than LadaGAN at $256 \times 256$ (2048M vs 18M), making it impractical for single-GPU training and highlighting LadaGAN's efficiency advantage.

\section{Conclusion}

In this paper, we presented LadaGAN, an efficient hybrid GAN architecture based on a linear additive-attention block called Ladaformer, which proved more suitable for both generator and discriminator than other efficient Transformer blocks, enabling stable GAN training. 
LadaGAN is gradient-stable and compares favorably with ConvNet and Transformer GANs on multiple benchmark datasets at different resolutions while requiring significantly fewer FLOPs. 
Moreover, compared with diffusion models and CT, LadaGAN achieves competitive performance at a fraction of the computational cost.

To the best of our knowledge, LadaGAN is among the first GANs to use linear additive attention in both generator and discriminator, providing evidence of their efficiency and expressive power and opening the door for future research on efficient GAN architectures. 
We believe LadaGAN can help laboratories perform experiments faster with limited computing budgets, enabling new breakthroughs while reducing energy consumption and carbon footprint.

As future work, we plan to extend LadaGAN to higher resolutions and more diverse datasets. 
Additionally, exploring advanced augmentation strategies (e.g., StyleGAN2-ADA, R3GAN) could further improve performance while maintaining LadaGAN's efficiency advantages.
Finally, we believe that the Ladaformer block and its compatibility with convolutions are worth exploring in other tasks, such as image and video classification.

\section*{Acknowledgements}
Emilio Morales-Juarez was supported by the National Council for Science and Technology (CONACYT), Mexico, scholarship number 782143.
This work was supported by DGAPA-UNAM through the PAPIIT program, project IN100624.

\bibliographystyle{splncs04}
\bibliography{references}

@techreport{krizhevsky2009learning,
  title={Learning multiple layers of features from tiny images},
  author={Krizhevsky, Alex},
  month = {April},
  year={2009},
  institution = {University of Toronto}
}

@article{kingma2014adam,
  title={Adam: A method for stochastic optimization},
  author={Kingma, Diederik P and Ba, Jimmy},
  journal={arXiv preprint arXiv:1412.6980},
  year={2014}
}

@inproceedings{goodfellow2014generative,
  title={Generative adversarial nets},
  author={Goodfellow, Ian and Pouget-Abadie, Jean and Mirza, Mehdi and Xu, Bing and Warde-Farley, David and Ozair, Sherjil and Courville, Aaron and Bengio, Yoshua},
  booktitle={Advances in neural information processing systems},
  pages={2672--2680},
  year={2014}
}

@article{bahdanau2014neural,
  title={Neural machine translation by jointly learning to align and translate},
  author={Bahdanau, Dzmitry and Cho, Kyunghyun and Bengio, Yoshua},
  journal={arXiv preprint arXiv:1409.0473},
  year={2014}
}

@article{yu15lsun,
    Author = {Yu, Fisher and Zhang, Yinda and Song, Shuran and Seff, Ari and Xiao, Jianxiong},
    Title = {LSUN: Construction of a Large-scale Image Dataset using Deep Learning with Humans in the Loop},
    Journal = {arXiv preprint arXiv:1506.03365},
    Year = {2015}
}

@inproceedings{liu2015faceattributes,
  title = {Deep Learning Face Attributes in the Wild},
  author = {Liu, Ziwei and Luo, Ping and Wang, Xiaogang and Tang, Xiaoou},
  booktitle = {Proceedings of International Conference on Computer Vision (ICCV)},
  month = {December},
  year = {2015} 
}

@inproceedings{ioffe2015batch,
  title={Batch normalization: Accelerating deep network training by reducing internal covariate shift},
  author={Ioffe, Sergey and Szegedy, Christian},
  booktitle={International conference on machine learning},
  pages={448--456},
  year={2015},
  organization={PMLR}
}

@InProceedings{arjovsky2017wasserstein,
  title = 	 {Wasserstein Generative Adversarial Networks},
  author = 	 {Martin Arjovsky and Soumith Chintala and L{\'e}on Bottou},
  booktitle = 	 {ICML},
  year = 	 {2017},
}

@inproceedings{gulrajani2017improved,
  title={Improved training of wasserstein gans},
  author={Gulrajani, Ishaan and Ahmed, Faruk and Arjovsky, Martin and Dumoulin, Vincent and Courville, Aaron C},
  booktitle={Advances in neural information processing systems},
  pages={5767--5777},
  year={2017}
}

@article{heusel2017gans,
  title={Gans trained by a two time-scale update rule converge to a local nash equilibrium},
  author={Heusel, Martin and Ramsauer, Hubert and Unterthiner, Thomas and Nessler, Bernhard and Hochreiter, Sepp},
  journal={Advances in neural information processing systems},
  volume={30},
  year={2017}
}

@article{vaswani2017attention,
  title={Attention is all you need},
  author={Vaswani, Ashish and Shazeer, Noam and Parmar, Niki and Uszkoreit, Jakob and Jones, Llion and Gomez, Aidan N and Kaiser, {\L}ukasz and Polosukhin, Illia},
  journal={Advances in neural information processing systems},
  volume={30},
  year={2017}
}

@inproceedings{
chen2018selfmod,
title={On Self Modulation for Generative Adversarial Networks},
author={Ting Chen and Mario Lucic and Neil Houlsby and Sylvain Gelly},
booktitle={ICLR},
year={2019},
url={https://openreview.net/forum?id=Hkl5aoR5tm},
}

@inproceedings{miyato2018spectral,
  title={Spectral normalization for generative adversarial networks},
  author={Miyato, Takeru and Kataoka, Toshiki and Koyama, Masanori and Yoshida, Yuichi},
  booktitle={ICLR},
  year={2018}
}

@inproceedings{brock2018large,
  title={Large scale gan training for high fidelity natural image synthesis},
  author={Brock, Andrew and Donahue, Jeff and Simonyan, Karen},
  booktitle={ICLR},
  year={2019}
}

@inproceedings{mescheder2018training,
  title={Which training methods for GANs do actually converge?},
  author={Mescheder, Lars and Geiger, Andreas and Nowozin, Sebastian},
  booktitle={International conference on machine learning},
  pages={3481--3490},
  year={2018},
  organization={PMLR}
}

@inproceedings{chen2019self,
  title={Self-supervised gans via auxiliary rotation loss},
  author={Chen, Ting and Zhai, Xiaohua and Ritter, Marvin and Lucic, Mario and Houlsby, Neil},
  booktitle={Proceedings of the IEEE/CVF conference on computer vision and pattern recognition},
  pages={12154--12163},
  year={2019}
}

@inproceedings{zhao2020differentiable,
  title={Differentiable Augmentation for Data-Efficient GAN Training},
  author={Zhao, Shengyu and Liu, Zhijian and Lin, Ji and Zhu, Jun-Yan and Han, Song},
  booktitle={NeurIPS},
  year={2020}
}

@inproceedings{karras2019style,
  title={A style-based generator architecture for generative adversarial networks},
  author={Karras, Tero and Laine, Samuli and Aila, Timo},
  booktitle={Proceedings of the IEEE conference on computer vision and pattern recognition},
  pages={4401--4410},
  year={2019}
}

@article{song2020denoising,
  title={Denoising diffusion implicit models},
  author={Song, Jiaming and Meng, Chenlin and Ermon, Stefano},
  journal={arXiv preprint arXiv:2010.02502},
  year={2020}
}

@article{ho2020denoising,
  title={Denoising diffusion probabilistic models},
  author={Ho, Jonathan and Jain, Ajay and Abbeel, Pieter},
  journal={Advances in Neural Information Processing Systems},
  volume={33},
  pages={6840--6851},
  year={2020}
}

@inproceedings{karras2020analyzing,
  title={Analyzing and improving the image quality of stylegan},
  author={Karras, Tero and Laine, Samuli and Aittala, Miika and Hellsten, Janne and Lehtinen, Jaakko and Aila, Timo},
  booktitle={Proceedings of the IEEE/CVF Conference on Computer Vision and Pattern Recognition},
  pages={8110--8119},
  year={2020}
}

@article{karras2020training,
  title={Training generative adversarial networks with limited data},
  author={Karras, Tero and Aittala, Miika and Hellsten, Janne and Laine, Samuli and Lehtinen, Jaakko and Aila, Timo},
  journal={arXiv preprint arXiv:2006.06676},
  year={2020}
}

@inproceedings{dosovitskiy2020image,
    title   = {An Image is Worth 16x16 Words: Transformers for Image Recognition at Scale},
    author  = {Alexey Dosovitskiy and Lucas Beyer and Alexander Kolesnikov and Dirk Weissenborn and Xiaohua Zhai and Thomas Unterthiner and Mostafa Dehghani and Matthias Minderer and Georg Heigold and Sylvain Gelly and Jakob Uszkoreit and Neil Houlsby},
    year    = {2021},
  booktitle = {ICLR},
}

@article{wu2021fastformer,
  title={Fastformer: Additive attention can be all you need},
  author={Wu, Chuhan and Wu, Fangzhao and Qi, Tao and Huang, Yongfeng and Xie, Xing},
  journal={arXiv preprint arXiv:2108.09084},
  year={2021}
}

@inproceedings{kim2021lipschitz,
  title={The lipschitz constant of self-attention},
  author={Kim, Hyunjik and Papamakarios, George and Mnih, Andriy},
  booktitle={International Conference on Machine Learning},
  pages={5562--5571},
  year={2021},
  organization={PMLR}
}

@inproceedings{liu2021towards,
  title={Towards faster and stabilized gan training for high-fidelity few-shot image synthesis},
  author={Liu, Bingchen and Zhu, Yizhe and Song, Kunpeng and Elgammal, Ahmed},
  booktitle={International Conference on Learning Representations},
  year={2020}
}

@article{jiang2021transgan,
  title={Transgan: Two transformers can make one strong gan},
  author={Jiang, Yifan and Chang, Shiyu and Wang, Zhangyang},
  journal={arXiv preprint arXiv:2102.07074},
  year={2021}
}

@article{zhao2021improved,
  title={Improved transformer for high-resolution gans},
  author={Zhao, Long and Zhang, Zizhao and Chen, Ting and Metaxas, Dimitris and Zhang, Han},
  journal={Advances in Neural Information Processing Systems},
  volume={34},
  year={2021}
}

@inproceedings{hudson2021gansformer,
  title={Generative adversarial transformers},
  author={Hudson, Drew A and Zitnick, Larry},
  booktitle={International conference on machine learning},
  pages={4487--4499},
  year={2021},
  organization={PMLR}
}

@inproceedings{lee2021vitgan,
  title={ViTGAN: Training GANs with Vision Transformers},
  author={Lee, Kwonjoon and Chang, Huiwen and Jiang, Lu and Zhang, Han and Tu, Zhuowen and Liu, Ce},
  booktitle={International Conference on Learning Representations},
  year={2021}
}

@inproceedings{zhang2021styleswin,
  title={Styleswin: Transformer-based gan for high-resolution image generation},
  author={Zhang, Bowen and Gu, Shuyang and Zhang, Bo and Bao, Jianmin and Chen, Dong and Wen, Fang and Wang, Yong and Guo, Baining},
  booktitle={Proceedings of the IEEE/CVF Conference on Computer Vision and Pattern Recognition},
  pages={11304--11314},
  year={2022}
}

@article{huang2024gan,
  title={The gan is dead; long live the gan! a modern gan baseline},
  author={Huang, Nick and Gokaslan, Aaron and Kuleshov, Volodymyr and Tompkin, James},
  journal={Advances in Neural Information Processing Systems},
  year={2024}
}

@article{dhariwal2021diffusion,
  title={Diffusion models beat gans on image synthesis},
  author={Dhariwal, Prafulla and Nichol, Alexander},
  journal={Advances in neural information processing systems},
  volume={34},
  pages={8780--8794},
  year={2021}
}

@inproceedings{zhai2022scaling,
  title={Scaling vision transformers},
  author={Zhai, Xiaohua and Kolesnikov, Alexander and Houlsby, Neil and Beyer, Lucas},
  booktitle={Proceedings of the IEEE/CVF Conference on Computer Vision and Pattern Recognition},
  pages={12104--12113},
  year={2022}
}

@article{ning2023input,
  title={Input Perturbation Reduces Exposure Bias in Diffusion Models},
  author={Ning, Mang and Sangineto, Enver and Porrello, Angelo and Calderara, Simone and Cucchiara, Rita},
  journal={arXiv preprint arXiv:2301.11706},
  year={2023}
}

@article{song2023consistency,
  title={Consistency Models},
  author={Song, Yang and Dhariwal, Prafulla and Chen, Mark and Sutskever, Ilya},
  journal={arXiv preprint arXiv:2303.01469},
  year={2023},
}

@inproceedings{nichol2021improved,
  title={Improved denoising diffusion probabilistic models},
  author={Nichol, Alexander Quinn and Dhariwal, Prafulla},
  booktitle={International Conference on Machine Learning},
  pages={8162--8171},
  year={2021},
  organization={PMLR}
}

\end{document}